\documentclass{article}
\usepackage{times}
\usepackage{algorithm,algorithmic}
\usepackage{amsmath,amssymb,amsthm}
\usepackage{graphicx}
\usepackage{comment,fullpage}
\usepackage{color}
\usepackage[round]{natbib}

\newtheorem{definition}{Definition}
\def\bfx{\mathbf{x}}
\def\bfy{\mathbf{y}}
\def\bfz{\mathbf{z}}
\def\bfu{\mathbf{u}}

\def\bfa{\mathbf{a}}
\def\bolda{\mathbf{a}}
\def\bfb{\mathbf{b}}
\def\bfg{\mathbf{g}}

\def\bmI{\mathbf{I}}
\def\bmK{\mathcal{K}}

\def\prox{\mathrm{prox}}
\def\diag{\mathrm{diag}}

\def\real{\mathbb{R}}

\def\K{\mathbf{K}}
\DeclareMathOperator*{\argmin}{arg\,min}

\def\bmH{\mathcal{H}}
\def\bmY{\mathcal{Y}}

\def\bmX{\mathcal{X}}

\def\bmL{\mathcal{L}}

\def\bmS{\mathcal{S}}
\def\bmG{\mathcal{G}}

\def\sdp{{\mathcal{S}^+_p}}

\newcommand{\norm}[1]{\left\Vert #1\right\Vert}

\newtheorem{theorem}{Theorem}

\title{Learning nonparametric differential equations with operator-valued kernels and gradient matching}
\author{Markus Heinonen$^{1,2}$, Florence d'Alch{\'e}-Buc$^{3,2}$\\
\small{
$^{1}$ IRSN, PRP-HOM/SRBE/L3R, 31 avenue de la Division Leclerc 92262 Fontenay-aux-roses, France}\\
\small{
$^{2}$ IBISC, Universit d'Evry, 23 BD de France, Evry 91037 cedex France}\\
\small{
$^{3}$ LTCI UMR cnrs 5141, Institut Mines-T{\'e}l{\'e}com, T{\'e}l{\'e}com ParisTech, 37 rue Dareau - 75014 Paris, France}\\
}

\begin{document}

\maketitle

\begin{abstract} Modeling dynamical systems with ordinary differential equations implies a mechanistic view of the process underlying the dynamics. However in many cases, this knowledge is not available. To overcome this issue, we introduce a general framework for nonparametric ODE models using penalized regression in Reproducing Kernel Hilbert Spaces (RKHS) based on operator-valued kernels. Moreover, we extend the scope of gradient matching approaches to nonparametric ODE. A smooth estimate of the solution ODE is built to provide an approximation of the derivative of the ODE solution which is in turn used to learn the nonparametric ODE model. This approach benefits from the flexibility of penalized regression in RKHS allowing for ridge or (structured) sparse regression as well. Very good results are shown on 3 different ODE systems.
\end{abstract}

\section{Introduction}

Dynamical systems modeling is a cornerstone of experimental sciences. In biology, as well as in physics and chemistry, modelers attempt to capture the dynamical behavior of a given system or a phenomenon in order to improve its understanding and eventually make predictions about its future state. Systems of coupled ordinary differential equations (ODEs) are undoubtedly the most widely used models in science. A single ordinary differential equation describes the change in rate of a given state variable as a function of other state variables. A set of coupled ODEs together with some initial condition can account for the full dynamics of a system. Even simple ODE functions can describe complex dynamical behaviours \citep{Hirsch04}. 

Ode modeling consists of the two tasks of (i) choosing the parametric ODE model, which is usually based on the knowledge of the system under study, and (ii) estimation of the parameters of the system from noisy observations. In many cases there is no obvious choice in favor of a given model. If many models are in competition, the choice can rely on hypothesis testing based model selection \citep{Cox1961,Vuong1989}. However it is not always easy to fulfill the assumptions of the statistical tests or propose a single mechanistic model. The aim of this work is to overcome these issuees by introducing a radically new angle for ODE modeling using nonparametric models, which sidestep the issue of model choice and provide a principled approach to parameter learning. A nonparametric model does not necessitate prior knowledge of the system at hand. 

More precisely, we consider the dynamics of a system governed by the following first-order multivariate ordinary differential equations:
\begin{eqnarray}
\dot{\bfx}(t) &=& f_{true}(\bfx(t)), \\
\bfx(t_0)     &=& x_0, \label{equadiff} 
\end{eqnarray}
where $\bfx(t) \in \mathbb{R}^p$ is the state vector of a $p$-dimensional dynamical system at time $t$, e.g. the ODE solution given the initial condition $\bfx_0$, and the $\dot{\bfx}(t) = \frac{d\bfx(t)}{dt}$ is a first order derivative of $\bfx(t)$ over time and $f_{true}$ is a vector-valued function. The ODE solution satisfies $$\bfx(t) = \bfx_0 + \int_0^t f_{true}(\bfx(\tau)) d\tau.$$ We assume that $f_{true}$ is unknown and we observe an $n$-length multivariate time series  $(\bfy_0 , \ldots , \bfy_{n-1})$ obtained from an additive noisy observation model at discrete time points $t_0, \ldots, t_{n-1}$:
\begin{equation}
\bfy_\ell = \bfx(t_\ell) + \epsilon_\ell,
\end{equation}
where $\epsilon_\ell$'s are i.i.d samples from a Gaussian distribution. 

In classical methods ODE approaches the parameters $\theta$ of the function $f(\bfx(t), \theta)$ are estimated with least squares approach, where the solution is simulated using a trial set of initial values for $\theta$, with subsequent parameter optimisation to maximise the simulated solution $\hat{x}_\theta(t_\ell) = \int_0^{t_\ell} f(\bfx(\tau), \theta) d\tau$ match against the observations $\bfy_\ell$. This approach, classically proposed off-the-shelf, involves computationally-intensive approximations and suffers from the scarcity of the observations. To overcome these issues, a family of methods proposed under different names of {\it collocation methods} \citep{Varah1982}, {\it gradient matching} \citep{Ellner2002} or {\it profiled estimation} \citep{Ramsay2007} have proposed to produce a smooth approximation $\hat{\bfx}(t)$ that can be used in turn as a surrogate $\dot{\hat{\bfx}}(t)$ of the true derivative $\dot{\bfx}(t)$ \citep{Bellman1971}. Then, we optimise the match between $\dot{\hat{\bfx}}$ and $f(\hat{\bfx},\theta)$, which does not require the costly integration step. In iterated estimation procedure the smoother and parameter estimation are iterated to correct the approximation errors in both terms. Using just two-steps has been analyzed in the parametric case in terms of asymptotics and it enjoys consistency results and provides nearly equal performance in the finite sample case \citep{Brunel2008,Gugushvili2011}.
 
In this work, we adopt the two-step gradient matching approach to the nonparametric ODE estimation to learn a nonparametric model $h \in \bmH$ to estimate the function $f_{true}$:
\begin{enumerate}
\item[\textbf{Step 1.}] Learn $g:\mathbb{R}\rightarrow \mathbb{R}^p$ in a functional space $\bmG$ to approximate the observed trajectory $\bfy_0, \ldots, \bfy_{n-1}$ at time points $t_0, \ldots, t_{n-1}$ with $t_0=0$.
\item[\textbf{Step 2.}] Given an approximation $g$ of the observed trajectory, learn $h:\mathbb{R}^p \rightarrow \mathbb{R}^p$ in a functional space $\bmH$ to approximate the differential $\dot{g}(t)$ with $h(g(t))$.
\end{enumerate}
The two steps play a very different role: with $g$, we want to capture the observed trajectories with enough accuracy so that $g(t)$ is close to the true value $x(t)$, and so that the derivative $\dot{g}(t)$ is close to the true differential $\dot{x}(t)$. When the covariance noise is assumed to be isotropic, each coordinate function $g_j; j=1, \ldots, p$ of vector-valued function $g$ can be learned separately, which we assume in this paper. In the second step, we want to discover dependencies between the vector spaces of state variables and state variable derivatives, and for that reason we will turn to vector-valued function approximation.  For that purpose, we propose as a key contribution to use the general framework of penalized regression for vector-valued functions in Reproducing Kernel Hilbert Spaces (RKHS) extending our preliminary works on the subject \citep{dalche2012, Lim2013a,Lim2013b}.  The RKHS theory offers a flexible framework for regularization in functional spaces defined by a positive definite kernel, with well-studied applications such  in scalar-valued function approximation, such as the SVM \citep{Aronszajn1950, Wahba90}.  Generalising the RKHS theory to vector-valued functions with operator-valued kernels \citep{Pedrick57,Senkene1973,Micchelli2005,Caponetto2008} has recently attracted a surge of attention for vector-valued functions approximation. Operator-valued kernels provide an elegant and powerful approach for multi-task regression \citep{Micchelli2005}, structured  supervised and semi-supervised output prediction \citep{Brouard2011,Dinuzzo2011}, functional regression \citep{Kadri2010} and nonparametric vector autoregression \citep{Lim2013a}. 

We introduce non-parametric operator-valued kernel-based regression models for gradient matching under $\ell_2$ penalties in Section \ref{sec:rkhs}. Section \ref{sec:multiple} extends the framework to learning from multiple time series obtained from multiple initial conditions. Section \ref{sec:sparse} proposes methods to introduce sparsity to the ODE models with $\ell_1$ and $\ell_1/\ell_2$ norms. Section \ref{sec:kernel} presents a method to learn the kernel function. Section \ref{sec:numerical} highlights the performance of the approach in two case studies of non-trivial classic ODE models and a realistic biological dataset with unknown model structure. We conclude in Section 7.

\section{Gradient Matching for nonparametric ODE }
\label{sec:rkhs}

We consider the following loss function based on empirical losses and penalty terms:
\begin{equation}
\label{eq:loss}
\bmL(h,g) = \frac{1}{2} \sum_{\ell=0}^{n-1} || g(t_\ell) - \bfy_\ell||^2 + \lambda_g ||g||^2 + \frac{1}{2} \sum_{\ell=1}^m || \dot{g}(t_\ell) - h(g(t_\ell)) ||^2 + \lambda_h ||h||^2.
\end{equation}

We present how {\bf Step 1} is solved using the classic tools of penalized regression in RKHS of scalar-valued functions and then introduce the tools based on operator-valued kernels devoted to vector-valued functions.

\subsection{Learning the smoother $g$}
\label{subset:g}

Various methods for data smoothing have been proposed in the literature and most of them (such as splines) can be described in the context of functional approximation in Reproducing Kernel Hilbert Spaces(see the seminal work of \citet{Wahba90} and also \cite{Pearce}). We apply standard kernel ridge regression as such an approximation.

For each state variable indexed by $j=1, \ldots, p$, we choose a positive definite kernel $k_j: \mathbb{R} \times \mathbb{R} \rightarrow \mathbb{R}$ and define the Hilbert space $\bmG_j$. We build a smoother $$g_j(t) = \sum_{i=0}^{n-1} b_{ij}k_j(t,t_i)$$ in the Hilbert space $\bmG_j$ by solving a kernel ridge regression problem. Given the observed data $\bmS_j = \{(t_0,y_{0j}), \ldots, (t_{n-1},y_{(n-1)j})\}$, minimizing the following loss: 
\begin{equation}\label{eq:gj}
\bmL(g_j; \bmS_j) = \frac{1}{2} \sum_{i=0}^{n-1} (y_{ij} - g_j(t_i))^2 + \lambda_{g,j} \norm{g_j}^2,
\end{equation}
with $||g_j||^2 = \mathbf{b}_j^T K_j \mathbf{b}_j$ leads to a unique minimizer:
\begin{equation}
\bfb_j = (K_j + \lambda_{g,j} Id)^{-1}\bfy_j,
\end{equation}
where $\bfb_j$ is the $n$-dimensional parameter vector of the solution model $g_j$ (for sake of simplicity, we avoid the hat notation), $K_j$ is the Gram kernel matrix computed on input data $t_0, \ldots, t_{n-1}$, Id is the $n \times n$ identity matrix, $\bfy_j$ is $n$-dimensional column vector of observed variable $j$. 

The derivatives $\dot{g}(t) = (\dot{g}_1(t), \ldots, \dot{g}_p(t))$ are straightforward to calculate as
\begin{equation}\label{eq:dotgj}
\dot{g}_j(t)= \sum_{i=0}^{n-1} b_{ji} \frac{d k_j(t,t_i)}{dt}.
\end{equation}

\subsection{Learning $h$ with operator-valued kernels}

Operator-valued kernel \citep{Alvarez2012} extends the well-known scalar-valued kernels in order to deal with functions with values in Hilbert Spaces. We present briefly the fundamentals of operator-valued kernels and the associated RKHS theory as introduced in \citep{Senkene1973, Micchelli2005}. Then we apply this theory to the case of functions with values in $\mathbb{R}^p$.

Let $\bmX$ be a non-empty set and $\bmY$ a Hilbert space. We note $\mathcal{L}(\bmY)$, the set
of all  bounded linear operators from  $\bmY$ to itself.  Given $M \in
\mathcal{L}(\bmY)$, $M^*$ denotes the adjoint of $M$. 

\begin{definition}[Operator-valued kernel]
  Let $\bmX$ be a non-empty set.  $K:\bmX
  \times \bmX \rightarrow \mathcal{L}(\bmY)$ is an operator-valued kernel if:
  \begin{itemize}
  \item $\forall  (x,x')   \in  \bmX  \times   \bmX$,  $K(x,x')  = K(x',x)^{*}$
  \item 
    $\forall m \in \mathbb{N}$, $\forall \mathcal{S}_m = \{(\bfx_i, \bfy_i)\}_{i=1}^m \subseteq
    \bmX \times \bmY$, 
    \begin{equation*}
      \sum_{i,j=1}^m \langle \bfy_i, K(\bfx_i,\bfx_j) \bfy_j\rangle_{\mathbb{R}^p}    \geq
      0.
    \end{equation*} 
  \end{itemize}
\end{definition}

Similarly to the scalar case, an operator-valued kernel allows to build a unique Reproducing Kernel Hilbert Space $\bmH_{K}$. First, the span of $\{h(\cdot) = \sum_{i} K(\cdot, x_i)\bfa_i,x_i \in \bmX \}$) is endowed with the following inner product: 
\begin{equation*}
\langle f,g \rangle_{\bmH_{K}} = \sum_{i,j} \langle \bfa_i,K(x_i,z_j)\bfb_j\rangle
\end{equation*}
with $f = \sum_i K(\cdot,x_i)\bfa_i$ and $g = \sum_j K(\cdot,z_j)\bfb_j$. This choices ensures the reproducing property: 
\begin{equation*}
\forall (x,\bfy,f) \in \bmX \times \bmY \times \bmH, \langle f, K(\cdot,x)\bfy \rangle_{\bmH_{K}} = \langle f(x), \bfy \rangle_{\bmY}
\end{equation*}
Then the corresponding norm  $\parallel \cdot \parallel_{\bmH_{0}}$ is defined by $\parallel f \parallel^2_{\bmH_{K}} = \langle f, f \rangle_{\bmH_{0}}$. Then $\bmH$ is completed by including the limits of Cauchy sequences for which the reproducing property still holds.
Approximation of a function in a RKHS enjoys representer theorems such as the following general one proved by \citet{Micchelli2005}
\begin{theorem}[\citet{Micchelli2005}]
\label{th:rep_theorem_general}
  Let $\bmX$ be a non-empty set, $\bmY$ a Hilbert Space and $K$an operator-valued kernel with values in $\bmL(\bmY)$. Let $\bmH$ the RKHS built from $K$. Let $\{(\bfx_\ell, \bfy_\ell) \in \bmX \times \bmY, \ell=1, \ldots, m\}$, a given set. Let $V:\bmY \times \bmY \rightarrow \mathbb{R}$ be a loss function, and $\lambda_h > 0$ a regularization parameter. Then any function inn $\bmH$ minimizing the following cost function:
  \begin{equation*}
  \mathcal{J}(h) =  \sum_{\ell=1}^m V(h(x_\ell),\bfy_\ell) + \lambda_h  \| h\|^2_{\bmH} \; ,
  \end{equation*}
admits an expansion:
  \begin{equation*}
    h(\cdot) = \sum_{\ell=1}^{m} K(\cdot,x_\ell) \bfa_\ell \; ,
  \end{equation*}
  where the coefficients $\bfa_\ell, \ell=1, \ldots, m$ are vectors in the Hilbert space $ \bmY$.
\end{theorem}

To solve the \textbf{Step 2} given $g$, we want to find a function 
\begin{align}
h(\bfx) &= \sum_{\ell=1}^{m} K(\bfx, g(t_\ell)) \bfa_\ell
\end{align}
intuitively minimizing the expected square gradient matching error over the time interval of interest $(0,T)$, plus a regularising term:
$$\bmL(h \,|\, g) = \frac{1}{2} \int_0^T || \dot{g}(t) - h(g(t))||^2 dt + \lambda_h ||h||^2.$$  \\
However, to apply the representer theorem to $h$, we are inclined to replace this expectation by an empirical mean (ignoring a factor) as
\begin{align}
\bmL(h \,|\, g, \tau_1, \ldots, \tau_m) &= \frac{1}{2} \sum_{\ell=1}^m || \dot{g}(\tau_\ell) - h(g(\tau_\ell)) ||^2 + \lambda_h ||h||^2
\end{align}
with $\tau_1, \ldots, \tau_m$ a sequence of $m$ positive reals, uniformly and independently sampled. We effectively use $m$ values along the trajectory estimate $g$ to act as the dataset to learn $h$ from, where usually $m \ge n$. 
Now, the representer theorem \ref{th:rep_theorem_general} applies to $h$ with the following choices: $\bmX = \bmY = \mathbb{R}^p$, $\bfx_\ell = g(\tau_\ell), \ell=1, \ldots, m$ and $\bfy_\ell = \dot{g}(\tau_\ell), \ell=1, \ldots, m$. This can be re-formulated as the Gradient Matching Representer Theorem:

\begin{theorem}[Gradient Representer Theorem]
\label{th:rep_theorem}
Let $g:\mathbb{R} \rightarrow \mathbb{R}^p$ be a vector-valued function, differentiable and $\tau_\ell, \ell=1, \ldots, m$, a set of positive reals. Let $K:\mathbb{R}^p \times \mathbb{R}^p \rightarrow \mathbb{R}$ a positive matrix-valued kernel and $\bmH_K=\bmH$ the RKHS built from $K$.
Let $V:\bmY \times \bmY \rightarrow \mathbb{R}$ be a loss function, and $\lambda_1 > 0$ a regularization parameter. Then any function of $\bmH$ minimizing the following cost function:
  \begin{equation*}
  \mathcal{J}(h) =  \sum_{\ell=1}^{m} V(h(g(\tau_\ell)),\dot{g}(\tau_\ell)) + \lambda_1  \| h\|^2_{\bmH},
  \end{equation*}
admits an expansion
 \begin{equation*}
   h(\cdot) = \sum_{\ell=1}^{m} K(\cdot,g(\tau_\ell)) \bfa_\ell,
 \end{equation*}
 where the coefficients $\bfa_\ell, \ell=1,  \ldots, m$ are vectors in $\mathbb{R}^p$.
\end{theorem}

\textbf{Step 2} is then solved as $\min_{h \in \bmH} \bmL(h \,|\, g, \tau_1, \ldots, \tau_m)$, given the kernel ridge regression solution $g(t)^T = (g_1(t), \ldots, g_p(t))$. The gradient matching problem admits a closed-form solution
\begin{equation*}
\label{eq:h-A-ridge}
\bfa = (\mathbf{K} + \lambda_h Id)^{-1} \dot{\mathbf{g}},
\end{equation*}
where the $pm$-dimensional $\dot{\mathbf{g}}$ is obtained by stacking the column vectors $\dot{g}(\tau_1), \ldots, \dot{g}(\tau_{m})$, and $\bfa$ is a $pm$-dimensional vector obtained by stacking the column vectors $\bfa_1,\ldots,\bfa_{m}$. $\mathbf{K}$ is a $m \times m$ block matrix, where each block is a matrix of size $p \times p$. The $(\ell,s)$'th block of $\K$ corresponds to the matrix $K(g(\tau_\ell),g(\tau_s))$.

\subsection{Operator-valued kernel families}

Several operator-valued kernels have been defined in the literature (\cite{Micchelli2005, Alvarez2012, Lim2013a}). We use two such kernels, which both are universal, SDP and based on the gaussian scalar kernel. First, the \emph{decomposable} kernel 
$$K_{DC}(\mathbf{x}, \mathbf{x}') = k( \mathbf{x}, \mathbf{x}') C,$$
where $k(\cdot, \cdot)$ is a standard gaussian scalar kernel, and $C \in \mathbb{R}^{p \times p}$ is a positive definite dependency structure matrix. Second, the \emph{transformable} kernel
$$K_{TF}(\mathbf{x}, \mathbf{x}')_{ij} = k( \mathbf{x}^i, \mathbf{x}'^j)$$ 
measures the pairwise similarities between features of data points, i.e. the variables between state vectors. Finally, we can also use the \emph{Hadamard} kernel
$$K_{had}(\mathbf{x}, \mathbf{x}') = K_{DC} \circ K_{TF} =   k( \mathbf{x}, \mathbf{x}') C \circ  K_{TF}( \mathbf{x}, \mathbf{x}').$$

\section{Learning from multiple initial conditions}
\label{sec:multiple}

In parametric ODEs, it is well known that using time series coming from different initial conditions reduces the non-identifiability of parameters. Similarly we also want to increase the accuracy of our estimate $h$ with multiple time series in the nonparametric case. However, in contrast to the parametric case, using $r$ multiple datasets produces $r$ models $h^i$ ($i=1,\ldots,r$). Given the assumption that there exists a true ODE model, it is supposed to be unique. We therefore propose to learn $r$ nonparametric models $h^{i}$ with a smoothness constraint that imposes that they should be close in terms of $\ell_2$ norm in the functional space $\bmH$, and hence they should give similar estimates for the same input. This corresponds to a multi-task approach \citep{evgeniou2004regularized} and is strongly related to the recent general framework of manifold regularization \citep{Quang2013}.

Let us assume that $r$ multivariate time series are observed, starting from $r$ different initial conditions. For simplicity, we assume that each time serie has the same length. {\bf Step 1} consists of learning $r$ vector-valued functions $g^{i}, i=1, \ldots, r$ from $\bmS_n^{i} = \{(t_0,\bfy_0^{i}), \ldots, (t_{n-1}, \bfy_{n-1}^{i}) \}$ as described in Section \ref{sec:rkhs}. The new loss to be minimized is 
\begin{align}
\bmL( h^1,\ldots,h^r | g^1,\ldots,g^r ) &= \sum_{i=1}^r \mathcal{L}(h^i | g^i) + \frac{1}{4} \lambda \sum_{i,j=1}^r ||h^i - h^j||^2  \\
  &= \frac{1}{2} \sum_{i=1}^r \left( || \dot{\mathbf{g}}^i - \mathbf{K}^{ii} \bfa^i ||^2 + \lambda_h {\bfa^i}^T \mathbf{K}^{ii} \bfa^i \right) + \frac{1}{2} \lambda_{sim} \left( r \sum_{i=1}^r {\bfa^i}^T \mathbf{K}^{ii} \bfa^i -  \sum_{i,j=1}^r {\bfa^i}^T  \mathbf{K}^{ij} \bfa^j \right)  \nonumber \\
    &= \frac{1}{2} \norm{\dot{\bfg} - \diag\bmK \bolda}^2 + \frac{1}{2} \lambda_h \bolda^T \diag\bmK \bolda + \frac{1}{2} \lambda_{sim}  \bolda^T (r \diag\bmK - \bmK) \bolda \nonumber,
  \end{align}
where $\dot{\mathbf{g}}^i$ is the stacked column vectors of all $\dot{g}(\tau_\ell)^i$'s, operator-valued kernel matrices $\mathbf{K}^{ij} = (K(g^i(\tau_j), g^j(\tau_k)))_{j,k=1}^m$ are the kernel matrices comparing time-series $i$ and $j$, and $\bfa^i$ is the stacked column vectors $(\bfa_1^i, \ldots, \bfa^i_m)$. Furthermore, the $\dot{\mathbf{g}}$ is concatenation of all $\dot{\mathbf{g}}^i$'s into a single column matrix, $\bfa$ is a concatenation of all $\bfa^i$'s, $\bmK = (\mathbf{K}^{ij})_{i,j=1}^r$ is a $r \times r$ block matrix of operator-valued kernel matrices $\mathbf{K}^{ij}$, and the diagonal $\diag\bmK$ has the diagonal blocks $\K^{ii}$ and zeroes elsewhere.

Similarly to standard kernel ridge regression and to semi-supervised kernel ridge regression, vector $\bolda$ can be obtained by annealing the gradient $\frac{\partial \bmL( h^1,\ldots,h^r | g^1,\ldots,g^r ) }{\partial \bolda}$ which gives the following closed form: 
\begin{align*}
\bolda  &= \left( \diag\bmK + (\lambda_h +r \lambda) Id - \lambda_{sim} (\diag\bmK)^{-1} \bmK \right)^{-1} \dot{\mathbf{g}},
\end{align*}
with $Id$ being here the identity matrix of dimension $mr$.
However to get an efficient approximation avoiding numerical issues, we use a stochastic averaged gradient descent in numerical experiments.

A single function can be constructed as the empirical average $\bar{h}(\cdot) = \frac{1}{r} \sum_{i=1}^{r} h^{i}(\cdot)$, representing the consensus model.

\section{Extension to sparse models}
\label{sec:sparse}

When learning the function $h$, we can use in principle as many training samples as we wish since the function $g$ and its analytical derivative $\dot{g}$ are available. The $\ell_{2}$ penalty associated with $h(\cdot)$ ensures the smoothness of the estimated functions. If we use the term \emph{support vectors} to refer to training vectors $g(\tau_{\ell})$ that have a non-zero contribution to the model, then an interesting goal is to try to use as few as possible training vectors and thus to try to reduce the number of corresponding non-zero parameters of $\bfa$. To achieve this, similarly to matrix-valued kernel-based autoregressive models \citep{Lim2013b}, we add to $\bmL( h | \hat{g})$ the following penalty $$\Omega_\bfa(\bfa) = \Omega_1(\bfa) + \Omega_{1,2}(\bfa)$$ composed of two sparsity-inducing terms:
\begin{align}
\Omega_1(\bfa) &= \alpha \norm{\bfa}_1 \\
\Omega_{1,2}(\bfa) &= (1-\alpha) \sum_{\ell=0}^N \norm{\bfa_{\ell}}_2
\end{align}
The first term $\Omega_1(\bfa)$ imposes the general lasso sparsity of the estimated function by aiming to set coefficients of the concatenated vector $\bfa$ to zero without taking into account the vector structure. The second term $\Omega_{1,2}$ imposes sparsity on the number of support vectors $\bfa_{\ell}$. The $\Omega_{1,2}$ is called mixed $\ell_1/\ell_2$-norm or the \emph{group lasso} \citep{yuan2006model}, which exhibits some interesting features: it behaves like an $\ell_1$-norm over the vectors $\bfa_\ell$ while within each vector $\bfa_\ell$, the coefficients are subject to an $\ell_2$-norm constraint.  The term $\alpha \in [0,1]$ gives a convex combination of $\ell_1$ and group lasso penalties ($\alpha=0$ gives group lasso, $\alpha=1$ gives $\ell_1$ penalty), while the $\lambda_h$ defines the overall regularisation effect. 
The new loss function $\bmL(\bfa) = \bmL( h | g) + \lambda_1 \Omega_{1,2}(\bfa)$ is still convex and can be decomposed into two terms: $\bmL_s({\bfa})$ which is smooth and differentiable with respect to $\bfa$ and $\bmL_{ns}(\bfa)$ which is non-smooth, but nevertheless convex and subdifferentiable with respect to $\bfa$:
\begin{equation}
\label{eq:loss_alpha}
\mathcal{L}(\bfa)= \bmL_s(\bfa) + \bmL_{ns}(\bfa), 
\end{equation}
where 
\begin{align*}
\bmL_s(\bfa) &= \frac{1}{2} \sum_{\ell=0}^{n-1} \norm{\dot{g}(\tau_{\ell}) - h(g(\tau_{\ell})) }^2_2 + \lambda_h ||h||_K^2  \\
\bmL_{ns}(\bfa) &=  \lambda_{1} \alpha ||\bfa||_1 +  \lambda_{1} (1 - \alpha) \sum_{\ell=0}^{n-1} \norm{\bfa_{\ell}}_2. \label{eq:l12}
\end{align*}

Recently, proximal gradient algorithms \citep{Combettes2011} have been proposed for solving problems of form \eqref{eq:loss_alpha} and shown to be successful in a number of learning tasks with non-smooth constraints. The main idea relies on using the proximal operator on the gradient term. For that purpose, we introduce the following notations: $L_{\bfa}$ is a Lipschitz constant -- the supremum -- of the gradient $\nabla_{\bfa} \bmL_s(\bfa)$. For $s>0$, the proximal operator of a function $\bmL_{ns}$ applied to some $\mathbf{v} \in \mathbb{R}^{np}$ is given by: 
\begin{equation*}
\text{prox}_{s}(\bmL_{ns})(\mathbf{v}) =  \argmin_\bfu \left\{ \bmL_{ns}(\bfu) + \frac{1}{2s} ||\bfu-\mathbf{v}||^2\right\}
\end{equation*}
The proximal gradient descent algorithm is presented in \ref{opt_K_}. Intermediary variables $t^{(m)}$ and $\bfa^{(m)}$ in Step 2 and Step 3 respectively are introduced to accelerate the proximal gradient method \citep{Beck2010}.

\begin{algorithm}[t]
\caption{Proximal gradient descent for minimizing (\ref{eq:loss_alpha})}
\label{opt_K_}
\begin{algorithmic}
\STATE \textbf{Inputs :} Initial solution $\bfa^{(0)} \in \real^{np}$, Lipschitz $L$
\STATE \textbf{Initialize :} $t^{(0)} = 1$
\STATE \textbf{until} convergence:
 \STATE \hspace{3mm} {\bf 1}: Compute $G = \nabla_\bfa \mathcal{L}_s(\bfa^{(m)})$ 
 \STATE \hspace{3mm} {\bf 2}: $\bfa^{(m)} = \prox_{\frac{\lambda}{L_\bfa},\alpha}(\ell_1 + \ell_1/\ell_2)\left( \bfa^{(m)} - \frac{1}{L_\bfa} G \right)$
 \STATE \hspace{3mm}  {\bf 3}: $t^{(m+1)} = \frac{1+\sqrt{1+4{t^{(m)}}^2}}{2}$
 \STATE \hspace{3mm}  {\bf 4}: $\bfa^{(m)} = \bfa^{(m)} + \frac{t^{(m)} - 1}{t^{(m+1)}}\left(\bfa^{(m)} - \bfa^{(m-1)}\right)$
\end{algorithmic}
\end{algorithm}

For a given vector $\mathbf{v} \in \mathbb{R}^{np}$, the proximal operator of $\ell_1$ is the element-wise shrinkage or soft-thresholding operator $\prox_{\mu \ell_1} : \mathbb{R}^{np} \rightarrow \mathbb{R}^{np}$
$$\prox_{\mu \ell_1}(\mathbf{u}) = \left(1-\frac{\mu}{||\mathbf{u}||_2}\right)_+ \mathbf{u},$$
while the proximal operator of the $\ell_1/\ell_2$ is the group-wise shrinkage operator $\prox_\mu(\ell_1/\ell_2) : \mathbb{R}^{np} \rightarrow \mathbb{R}^{np}$
$$\prox_{\mu \ell_1/\ell_2}(\mathbf{u})_{\mathcal{I}} = \left(1-\frac{\mu}{||\mathbf{u}_{\bmI}||_2}\right)_+ \mathbf{u}_{\bmI},$$
where $\mathbf{u}_{\mathcal{I}}$ denotes the coefficients of $\mathbf{u}$ indexed by  $\bmI$. $\bmI$ denotes the group indexes. The proximal operator of the combined $\ell_1$ and $\ell_1/\ell_2$ regularisers  is the so called `sparse group lasso', and it's defined as
$$\prox_{\lambda,\alpha \ell_1 + \ell_1/\ell_2}(\mathbf{u}) =  \prox_{\lambda \alpha}(\ell_1/\ell_2)\left( \prox_{\lambda (1-\alpha)}(\ell_1)(\mathbf{u})\right) $$

We initialise the proximal algorithm with the solution on the smooth part $\mathcal{L}_s(\bfa)$. The gradient of the smooth part is $\nabla_\bfa \mathcal{L}_s(\bfa) = (\K + \lambda_h Id) \bfa + \dot{\mathbf{g}}$. By definition
\begin{align*}
||\nabla \mathcal{L}_s(\bfa) - \nabla \mathcal{L}_s(\bfa')|| \le || \K + \lambda_h I ||_F \cdot  ||\bfa - \bfa'||
\end{align*}
giving a Lipschitz constant $L_\bfa = || \K + \lambda_h Id ||_F$.

\section{Kernel learning}
\label{sec:kernel}

In order to build the vector-valued function $h$, various matrix-valued kernels can be chosen either among those already described in the literature (see for instance \citet{Caponetto2008} and \citet{Alvarez2012}) or built using the closure properties of operator-valued kernels as in \citet{Lim2013a}. Here, as an example, we choose the decomposable kernel which was originally proposed to tackle multi-task learning problems in \cite{Micchelli2005}. In the case of matrix-valued kernel with values in $\bmL(\mathbb{R}^p)$, the decomposable kernel $K: \mathbb{R}^p \times \mathbb{R}^p \rightarrow \bmL(\mathbb{R}^p)$ defined as follows:
\begin{equation}
\forall (\bfx,\bfz), K(\bfx,\bfz) = C k(\bfx,\bfz),
\end{equation}
where C is a $p \times p$ positive semi-definite matrix.\\

The non-zero elements $C_{ij}$ reflect dependency relationship between variables. Hence, we desire to learn the matrix $C$ of the decomposable kernel $K(x,x') = C \cdot k(x,x')$ of the ode function $h(g(t)) = \sum_{\ell=1}^N K_C(g(t), \bfy_\ell) \bfa_\ell$. We resort to two-step optimization where we alternatively optimize (i) the loss function \eqref{eq:lossC} given a constant $C$, and (ii) the loss function given a constant $\bfa$ and $\bfb$:
\begin{align}
\label{eq:lossC}
\bmL(C) &= \underbrace{\frac{1}{2} \sum_{\ell=1}^{m} || \dot{g}(t_\ell) - h_C(g(t_\ell)) ||^2 + \lambda_h ||h||^2}_{\mathcal{L}_s(C)} + \mathbf{1}_\sdp(C),
\end{align}
where $\sdp$ is the cone of positive semidefinite $p \times p$ matrices, and $\mathbf{1}_\sdp$ denotes an indicator function with $\mathbf{1}_\sdp(C) = 0$ if $C \in \sdp$ and $\infty$ otherwise. 

The first two terms $\mathcal{L}_s$ of the loss function are differentiable with respect to $C$, while the remaining terms are non-smooth, but convex and sub-differentiable with respect to $C$. To optimise the loss function, we employ proximal algorithms, which here reduces to projected gradient descent where after each iteration we project the value of $C$ to the cone of SDP matrices \citep{Richard2012}.  Our algorithm is presented in Algorithm \ref{alg:proxC}.

\begin{algorithm}[t]
\caption{Incremental proximal gradient descent for minimizing (\ref{eq:lossC})}  
\label{alg:proxC}
\begin{algorithmic}
\STATE \textbf{Inputs:} Initial solution $C_0 \in \mathbb{R}^{p \times p}$, Lipschitz $L_C$ 
\STATE \textbf{Initialize:} $C = C_0$ 
\STATE {\bf until} convergence: 
\STATE \hspace{3mm} {\bf Step 1}: $C = C - \frac{1}{L_C} \nabla_C \mathcal{L}_s(C)$ 
\STATE \hspace{3mm} {\bf Step 2}: $C = \prox_{\sdp}\left(C \right)$ 
\end{algorithmic}
\end{algorithm}

The projection onto the cone of positive semidefinite matrices is
$$\prox_{\sdp}(C) = \argmin_{X \in \sdp} || X - C||_F = U \Lambda_+ U^T,$$
where $U$ is the orthonormal eigenvectors of $C$ and $\Lambda_+ = \max(\Lambda,0)$ is the non-negative eigenvalue matrix of the corresponding eigenvalues $\Lambda$.

The gradient of the smooth part of the loss function is
\begin{align*}
\nabla_C \mathcal{L}_s(C) &= - EKA^T + AKA^T
\end{align*}
where $K = (k(g(t_\ell), g(t_s)))_{\ell,s=1}^{m}$ is a scalar kernel matrix, $E = (\dot{g}(t_\ell) - h(g(t_\ell))_{\ell=1}^{m}$ is a matrix of differences and $A = (\bfa_1, \ldots, \bfa_m) \in \mathbb{R}^{p \times m}$ is a matrix with $\bfa_\ell$ as columns. Due to the SDP constraint, the gradient matrix is symmetric.

The Lipschitz constant $L_C = ||H_I K A^T||_F^2$ is due to the property
\begin{align*}
|| \nabla_C \mathcal{L}_s(C) - \nabla_C \mathcal{L}_s(C')||^2_F \le ||C - C'||_F^2 \cdot || H_I K A^T ||_F^2,
\end{align*}
where $H_I = (h_I(g(t_\ell)))_{\ell=1}^{m}$ is the matrix of predictions from $h(g(t))$. We have the linear property $H_C = C H_I$.

\section{Numerical results}
\label{sec:numerical}

We perform numerical experiments on three datasets to highlight the performance of the novel OKODE framework in ODE estimation, analyze the resulting models, and finally compare our approach to the state-of-the-art ODE estimation methods.

\subsection{Experimental setting}

Throughout the experiments we use independent kernel ridge regression models for the variables of $g_i(t)$ with a gaussian scalar kernel $k_i(x,x') = \exp(-\gamma_i ||x - x'||^2)$ with hyperparameter $\lambda_i$. For the learning of $h$ we use a decomposable kernel $K(g(t), g(t_{\ell})) = C \cdot k(g(t), g(t_\ell))$ with a gaussian scalar kernel $k(x,x') = \exp(-\gamma ||x - x'||^2)$. The different errors of the smoother $g$, the ode $h$ and the trajectory $\int h(g(t)) dt$ are summarized as
\begin{align*}
\text{Smoothing error:   } & \sum_{\ell=0}^{n-1} || \bfy_{\ell} - g(t_{\ell}) ||^2 \\
\text{Gradient matching error:   } & \sum_{\ell=1}^{m} || \dot{g}(t_\ell) - h(g(t_\ell))||^2 \\
\text{Trajectory error:   } & \sum_{\ell=1}^m || \bfy_{\ell} - \int_{0}^{t_{\ell}} h(g(\tau)) d\tau ||^2.
\end{align*}
We choose the hyperparameters $(\gamma_i, \lambda_i)$ using leave-one-out cross-validation over the dataset $\mathcal{S} = (t_i, \bfy_i)_{i=1}^n$ to minimize the empirical smoothing error. For the $g$, we choose the hyperparameters $(\gamma,\lambda)$ through a grid search by minimizing the empirical trajectory error of the resulting model against the observations $\mathcal{S}$. We learn the optimal matrix $C$ using iterative proximal descent. We solve the multiple time series model with stochastic gradient descent with batches of 10 coefficients with averaging after the first epoch, for a total of 20 epochs. For multiple time series, we set manually $\lambda=0.1$. 

\subsection{ODE estimation}

We apply OKODE on two dynamic models with true mechanics known, and on dataset specifically designed to represent a realistic noisy biological time series with no known underlying model. 


\subsubsection{Fitz-Hugh Nagumo model}
\label{sec:fhn}

The FitzHugh-Nagumo equations \citep{Fitzhugh1961,Nagumo1962} approximate the spike potentials in the giant axon of squid neurons with a model
\begin{align*}
\dot{V} &= c(V - V^3/3 + R) \\
\dot{R} &= -1/c (V - a + bR).
\end{align*}
The model describes the dependency between the voltage $V$ across the axon membrane and the recovery variable $R$ of the outward currents. We assign true values $(0.2, 0.2, 3)$ for the parameters $(a,b,c)$. The FHN model is relatively simple non-linear dynamical system, however it has proven challenging to learn with numerous local optima \cite{Ramsay2007}. We sample $n=41$ observations over regularly spaced time points with added isotropic, zero-mean Gaussian noise with $\sigma^2 = 0.1$.

\begin{figure}[h!]
 \label{fig:fhn}
 \begin{center}
  \includegraphics[width=4.2in]{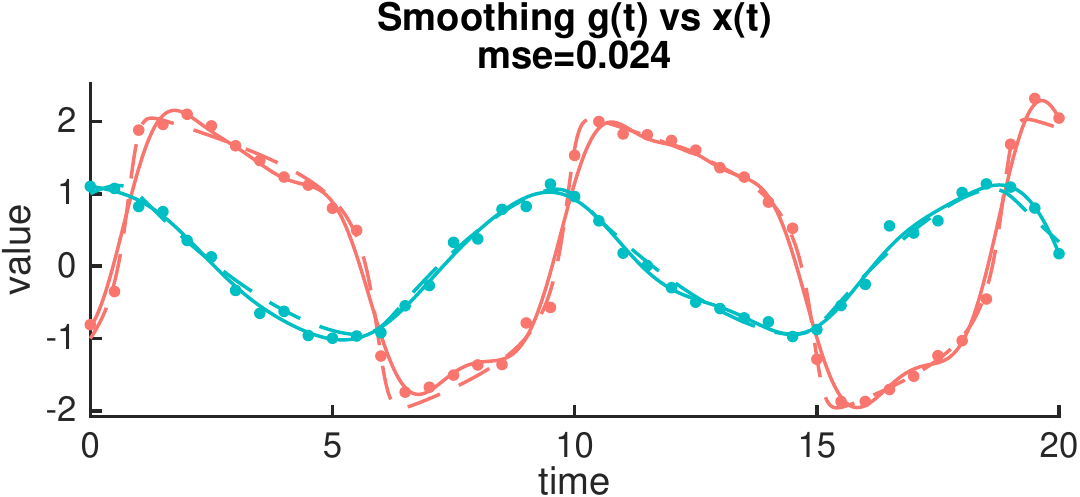}  \\
  \includegraphics[width=4.2in]{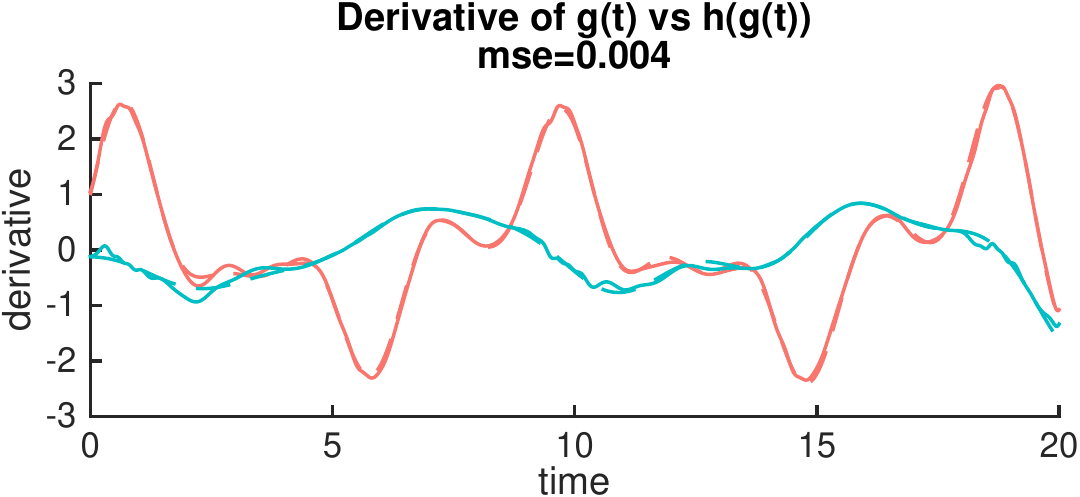} \\
  \includegraphics[width=4.2in]{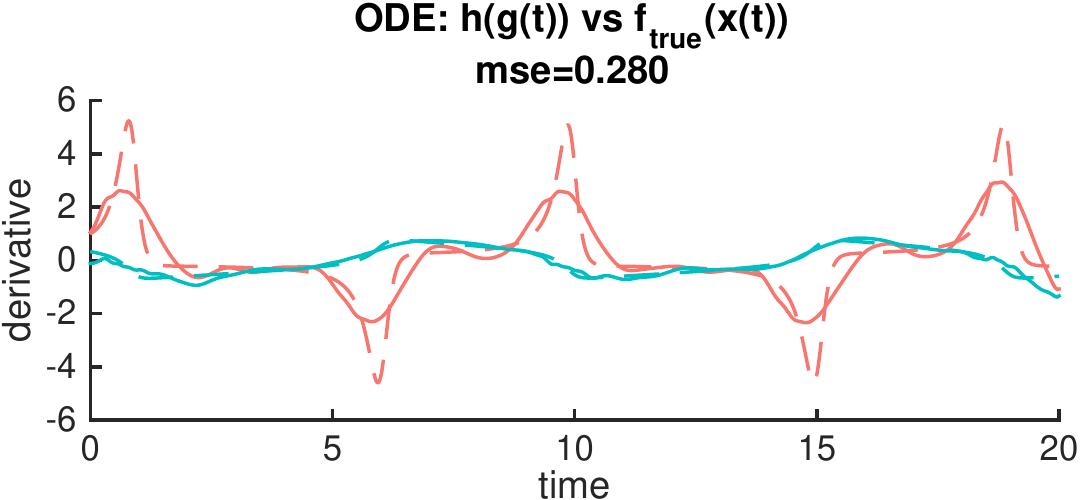} \\
  \includegraphics[width=4.2in]{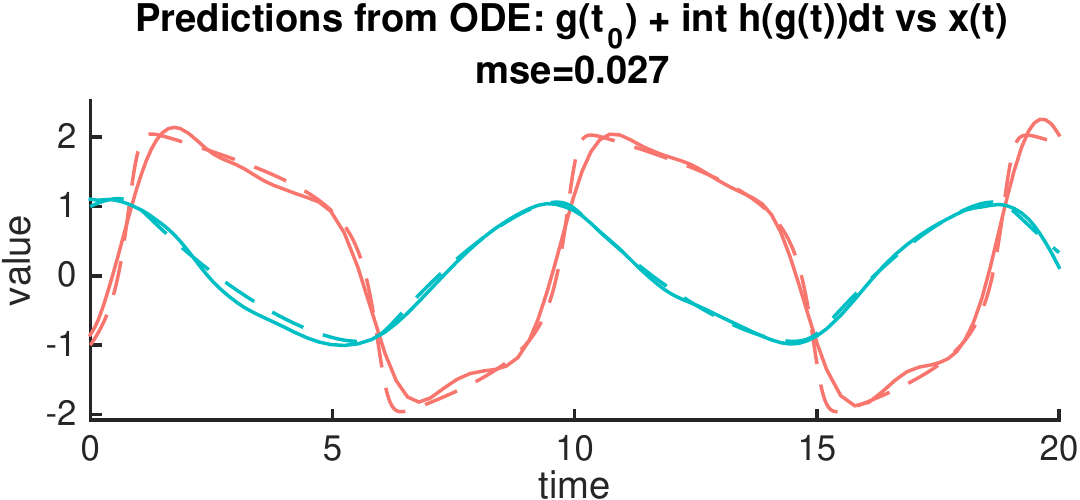}
 \end{center}
 \caption{Learned OKODE of the FHN model.}
\end{figure}

The Figure \ref{fig:fhn} presents the learned OKODE model with $m=101$ parameter vectors. The four figures depict from top to bottom the (i) smoother $g$, (ii) the gradient $h(g(t))$ against $\dot{g}(t)$, (iii) the $h$ against the $f_true$, and (iv) the estimated trajectory from $h$. The estimated trajectory matches the true well, but tends to underestimate the derivatives around the sharp turns of the red curve.

\subsubsection{Calcium model}

The calcium model \citep{peifer07} represents the oscillations of calcium signaling in eucaryotic cells by modeling the concentrations of free calcium in cytoplasm $Ca_{cyt}$ and in the endoplasmic reticulum $Ca_{re}$, and active $G_\alpha$ and phospholipase-C, $PLP_C$. The system consists of 17 parameters. We sample $n=67$ regularly spaced observations with added zero-truncated Gaussian noise $\mathcal{N}(0,0.1)$. We use $m=101$ parameter vectors.
 17 parameters determine the system
\begin{align*}
\dot{G_\alpha} &= k_1 + k_2 G_\alpha - k_3 P_C R_1(G_\alpha) - k_4 Ca_{c} R_2(G_\alpha) \\
\dot{P}_C &= k_5 G_\alpha - k_6 R_3(P_C) \\
\dot{Ca}_{c} &= k_7 P_C Ca_{c} R_4(Ca_{r}) + k_8 P_C + k_9 G_\alpha - k_{10} R_5(Ca_{c}) - k_{11} R_6(Ca_{c}) \\
\dot{Ca}_{r} &= -k_7 P_C Ca_{c} R_4(Ca_{r}) + k_11 R_6(Ca_{c}). 
\end{align*}
where $R_i(x) = \frac{x}{x + Km_i}$. We sample $n=67$ regularly spaced observations with added zero-truncated Gaussian noise $\mathcal{N}(0,0.1)$. We use $m=101$ parameter vectors.

\begin{figure}[h!]
 \begin{center}
  \includegraphics[width=0.5\columnwidth]{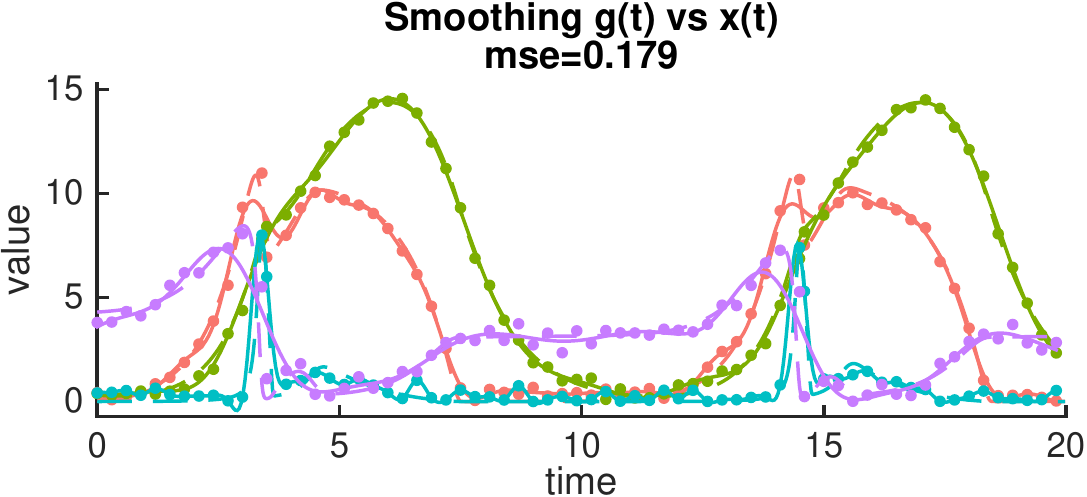} \\
  \includegraphics[width=0.5\columnwidth]{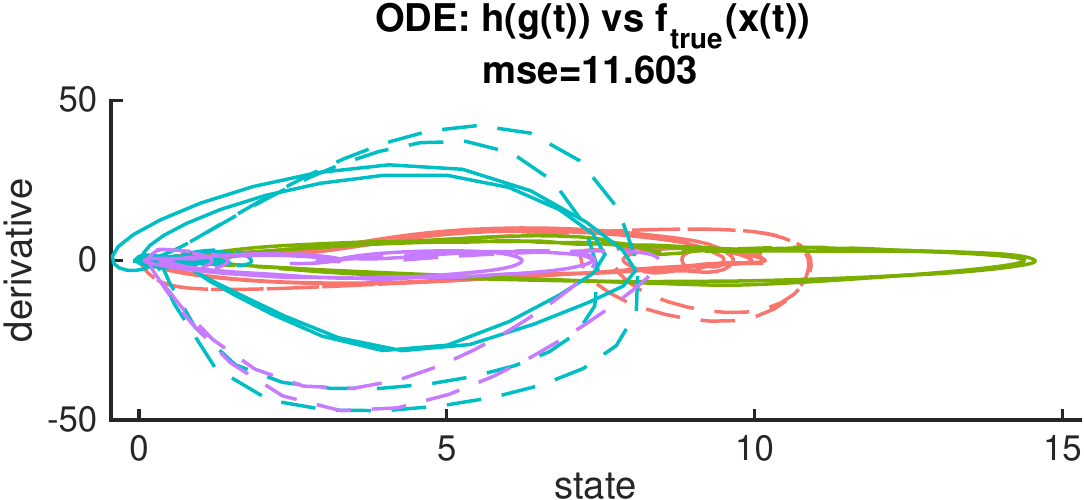} \\
  \includegraphics[width=0.5\columnwidth]{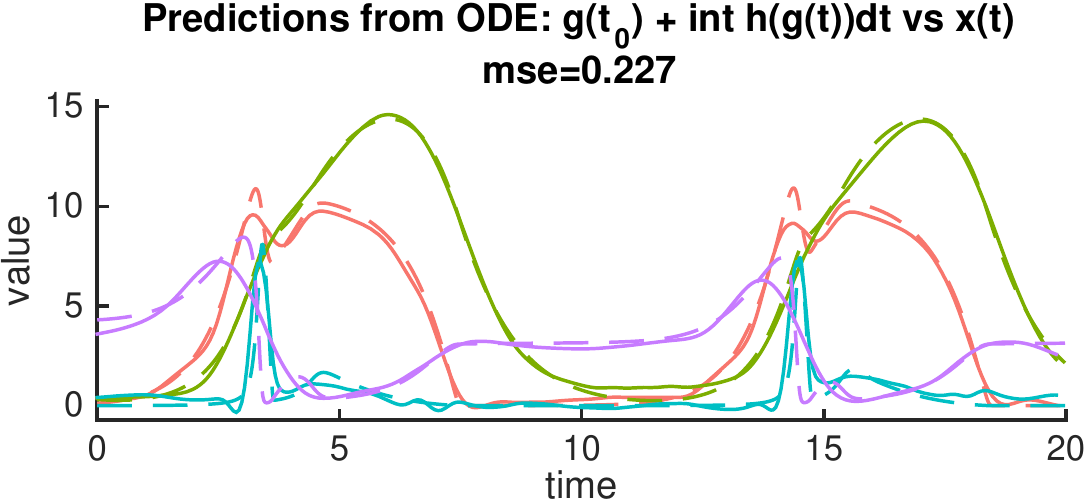} 
 \end{center}
 \label{fig:calcium}
 \caption{Learned OKODE of the calcium model.}
\end{figure}

The Figure \ref{fig:calcium} depicts the (i) smoother $g$, (ii) the estimated $h(x)$ and $f_{true}(x)$ over $x$ and (iii) finally the predicted trajectory. The smoother doesn't learn the peaks of the red curve, which is reflected in the estimated trajectory.

\subsubsection{DREAM3}

The DREAM3 dataset consists of realistically simulated biological time-series of 10 variables with $n=21$ noisy observations. The corresponding ODE models are deliberately unknown. We employ the DREAM3 dataset as a representative of a biological, noisy dataset with no gold standard to perform exploratory ODE modeling. We note that such data has not been applicaple to ODE analysis. 

\begin{figure}[h!]
 \begin{center}
  \includegraphics[width=0.6\columnwidth]{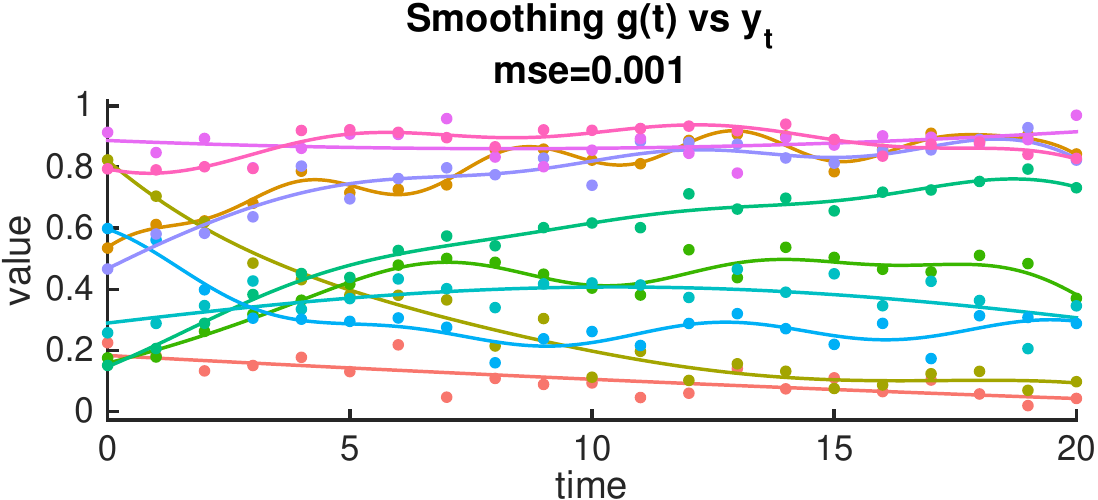} \\
  \includegraphics[width=0.6\columnwidth]{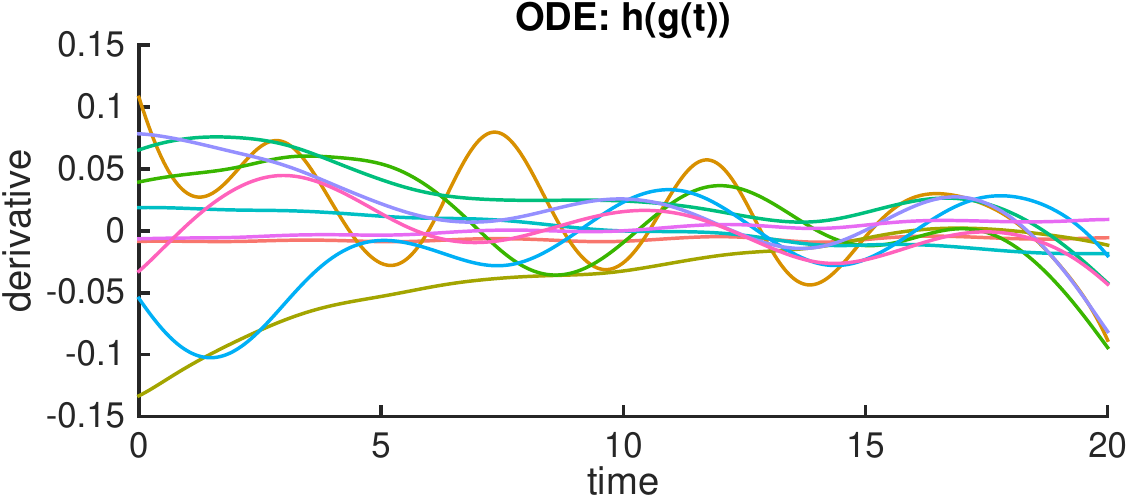} \\
  \includegraphics[width=0.6\columnwidth]{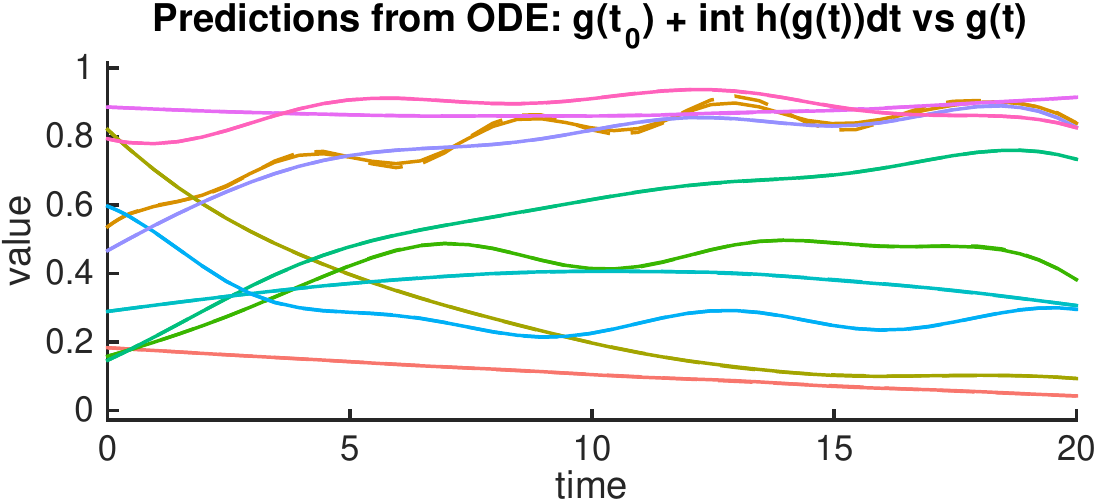} 
 \end{center}
 \caption{Learned OKODE of the DREAM3 model.}
 \label{fig:dream3}
\end{figure}

The Figure \ref{fig:dream3} shows the result of applying the proposed method to DREAM3 dataset with $m=101$. The ODE model reconstructs the smoother $g$ rather well, but there is considerable uncertainty in how smooth the function $g$ should be. 

\subsection{Model sparsity}

\begin{figure}[h!]
 \begin{center}
  \includegraphics[width=0.65\columnwidth]{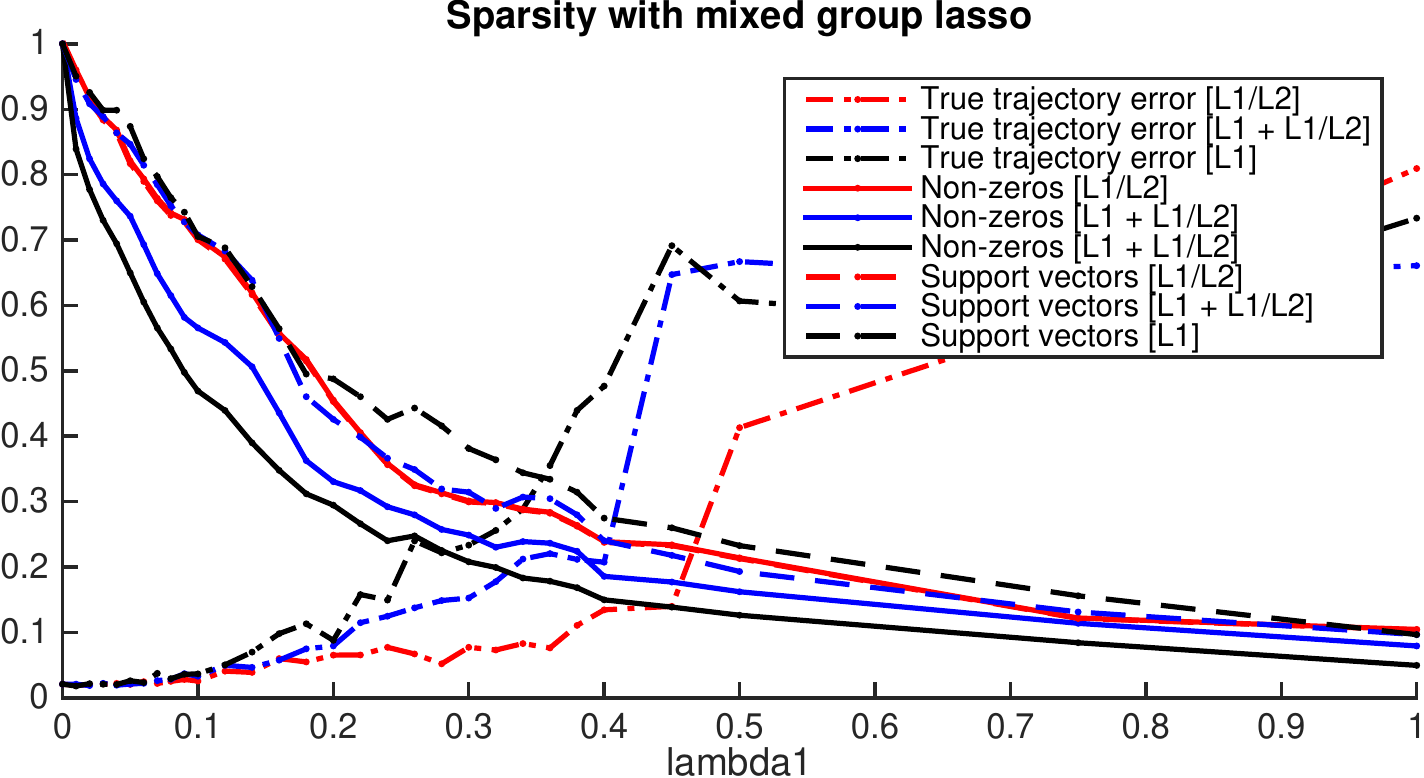} 
 \end{center}
 \caption{The tradeoff between trajectory error and sparsity under sparse norms over the FHN dataset.}
 \label{fig:sparsity}
\end{figure}

Figure \ref{fig:sparsity} indicates true trajectory errors and levels of sparsity when learning the FHN model with different values of $\alpha$ and with a high number of $m=404$ model points and $n=41$ data points as in Section \ref{sec:fhn}. Approximately half of the coefficients can be set to zero without large effect on the trajectory error. However, as an automated approach for selection of $m$, using sparsity is quite unrobust.

\subsection{Multiple initial conditions}

\begin{figure}[h!]
 \begin{center}
  \includegraphics[width=0.48\columnwidth]{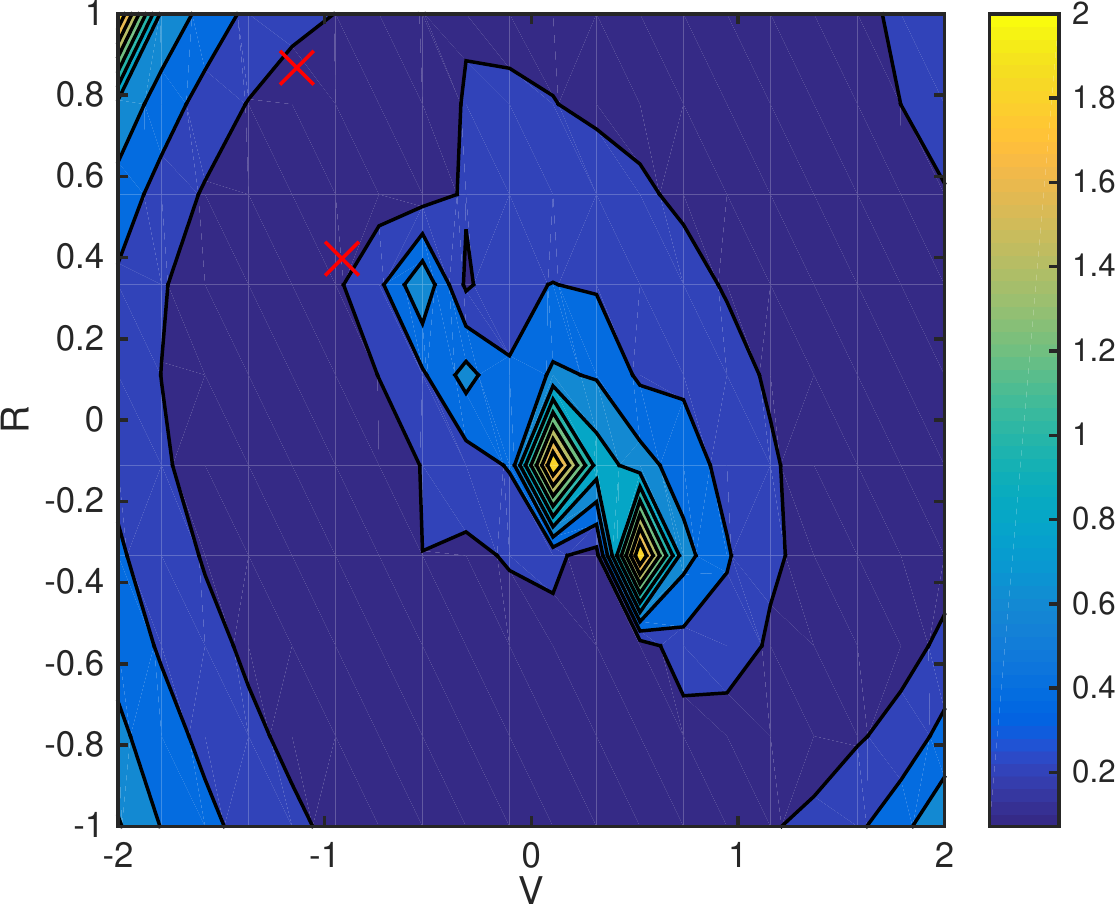} 
  \includegraphics[width=0.48\columnwidth]{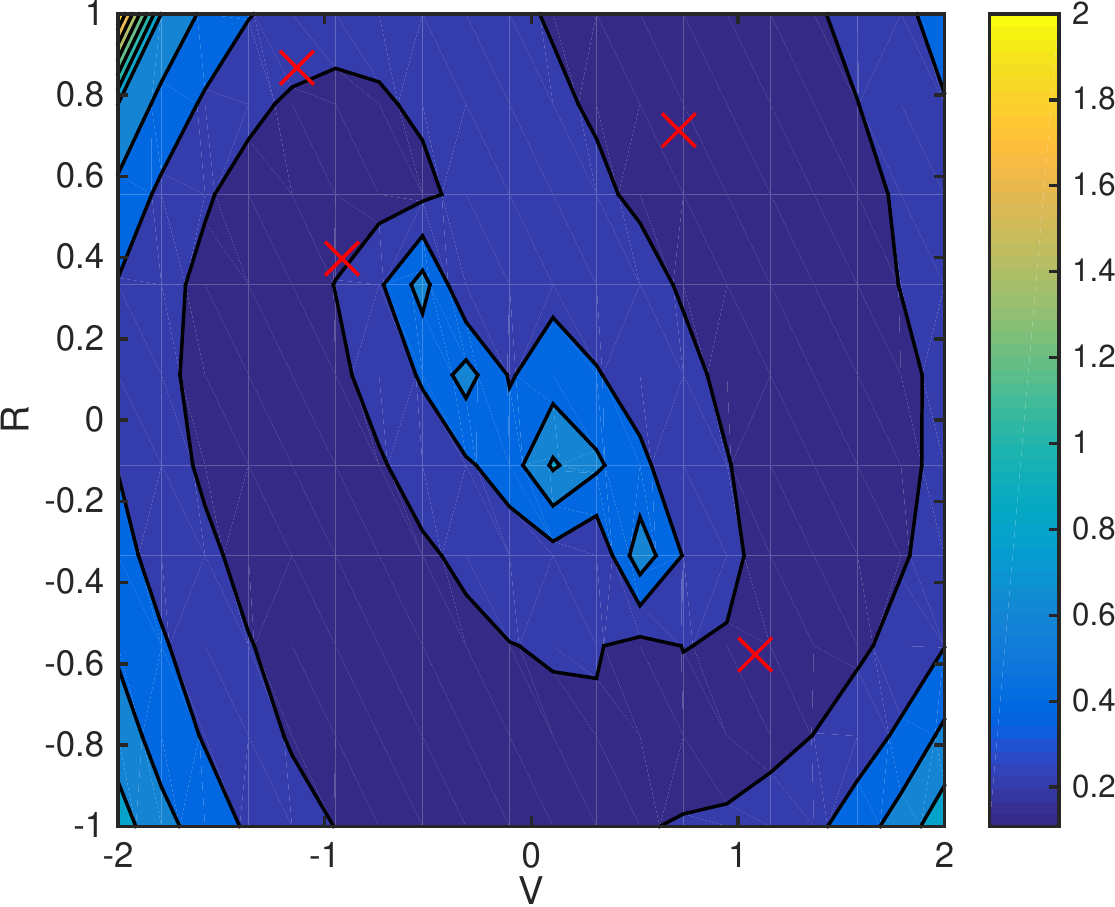}
 \end{center}
 \caption{The generalisation error of the ODE model with a model learned from 2 or 4 time series. The red crosses indicate the initial conditions of the time-series the model was learned from.}
 \label{fig:errmap}
\end{figure}

We study learning a nonparametric model using multiple time-series on the FHN dataset. We are interested in the ability of the model to estimate accurate trajectories from arbitrary initial conditions. In Figure \ref{fig:errmap} we plot the true trajectory error of a model learned with 2 or 4 time series over the space of $(V,R)$ values as initial values. As expected, adding more time-series improves the model's ability to generalise, and provides a more accurate ODE model.

\subsection{Comparison against parametric estimation}

We compare our non-parametric OKODE framework against the iterative method of \cite{Ramsay2007}, and to the classic parametric parameter estimation on the FHN model with initial values $(-1,1)$. The parametric approach has three parameters $(a,b,c)$ to learn, and is expected to perform well as the ODE model is known. For a more realistic comparison, we also estimate the parameters when we assume only a third order model with $14$ parameters
\begin{align*}
\dot{V} &= p_1 V + p_2 V^2 + p_3 V^3 + p_4 R + p_5 R^2 + p_6 R^3 + p_7 \\
\dot{R} &= p_8 V + p_9 V^2 + p_{10} V^3 + p_{11} R + p_{12} R^2 + p_{13} R^3 + p_{14}
\end{align*}
where the true values are $p_1 = 3$, $p_3 = -1$, $p_4 = 3$, $p_8 = -1/3$, $p_{11} = p_{14} = -1/6$ and remaining values are zero. We use MATLAB's \texttt{fminsearch} and do $100$ restarts from random initial values from $\mathcal{N}(0,1)$ to estimate the best values. Table \ref{table:comparison} highlights the true trajectory errors of the ODE. As expected, given a known ODE model, the parametric model achieves a near perfect performance. The method of Ramsay and OKODE perform well, while the parametric solver fails if the ODE model is less rigorously specified.

\begin{table}[h!]
\caption{Mean square errors of ODE methods on the FHN dataset.}
\begin{center}
\begin{tabular}{|l|r|}
\hline
Method                 & MSE \\
\hline          
Ramsay                  & $0.0094$ \\
Parametric, 3 free coefficients  & $0.0001$ \\
Parametric, 14 free coefficients & $0.4925$ \\
OKODE                   & $0.0270$ \\
\hline
\end{tabular}
\end{center}
\label{table:comparison}
\end{table}

\section{Conclusion}
We described a new framework for nonparametric ODE modeling and estimation. We showed that matrix-valued kernel-based regression were especially well appropriate to build estimates in a two-step gradient matching approach. The flexibility of penalized regression in RKHS provides a way to address realistic tasks in ODE estimation such as learning from multiple initial conditions.
We show that these models can be learned as well using nonsmooth constraints with the help of proximal gradient algorithms and also discuss the relevance of sparse models. 
Future works concerns the study of this approach in presence of heteroscedastic noise. A Bayesian view of this approach will also be of interest (\cite{Calderhead2009, Dondelinger2013}). Finally, another important issue is the scaling up of these methods to large dynamical systems.
\section{Acknowledgements}
This work was supported by ANR-09-SYSC-009 and Electricit{\'e} de France (Groupe Gestion Projet-Radioprotection) and Institut de Radioprotection et de S\^uret{\'e} nucl{\'e}aire (programme ROSIRIS).
\bibliographystyle{plainnat}
\bibliography{okode}

\begin{thebibliography}{34}
\providecommand{\natexlab}[1]{#1}
\providecommand{\url}[1]{\texttt{#1}}
\expandafter\ifx\csname urlstyle\endcsname\relax
  \providecommand{\doi}[1]{doi: #1}\else
  \providecommand{\doi}{doi: \begingroup \urlstyle{rm}\Url}\fi

\bibitem[{\'A}lvarez et~al.(2012){\'A}lvarez, Rosasco, and
  Lawrence]{Alvarez2012}
M.~A. {\'A}lvarez, L.~Rosasco, and N.~D. Lawrence.
\newblock Kernels for vector-valued functions: a review.
\newblock \emph{Foundations and Trends in Machine Learning}, 4\penalty0
  (3):\penalty0 195--266, 2012.

\bibitem[Aronszajn(1950)]{Aronszajn1950}
N.~Aronszajn.
\newblock Theory of reproducing kernels.
\newblock \emph{Transactions of the American mathematical society}, pages
  337--404, 1950.

\bibitem[Beck and Teboulle(2010)]{Beck2010}
A.~Beck and M.~Teboulle.
\newblock Gradient-based algorithms with applications to signal recovery
  problems.
\newblock In DP~Palomar and YC~Eldar, editors, \emph{Convex Optimization in
  Signal Processing and Communications}, pages 42--88. Cambridge press, 2010.

\bibitem[Bellman and Roth(1971)]{Bellman1971}
R.~Bellman and R.~Roth.
\newblock The use of splines with unknown end points in the identification of
  systems.
\newblock \emph{J. Math. Anal. Appl.}, 34:\penalty0 26--33, 1971.

\bibitem[Brouard et~al.(2011)Brouard, d'Alch{\'e} Buc, and
  Szafranski]{Brouard2011}
C.~Brouard, F.~d'Alch{\'e} Buc, and M.~Szafranski.
\newblock Semi-supervised penalized output kernel regression for link
  prediction.
\newblock In \emph{Proc. of the 28th Int. Conf. on Machine Learning (ICML-11)},
  pages 593--600, 2011.

\bibitem[Brunel(2008)]{Brunel2008}
N.~J-B. Brunel.
\newblock Parameter estimation of ode's via nonparametric estimators.
\newblock \emph{Electronic Journal of Statistics}, 2:\penalty0 1242--1267,
  2008.

\bibitem[Calderhead et~al.(2009)Calderhead, Girolami, and
  Lawrence]{Calderhead2009}
B~Calderhead, M~Girolami, and N~Lawrence.
\newblock Accelerating bayesian inference over nonlinear differential equations
  with gaussian processes.
\newblock In \emph{NIPS}, 2009.

\bibitem[Caponnetto et~al.(2008)Caponnetto, Micchelli, Pontil, and
  Ying]{Caponetto2008}
A.~Caponnetto, C.~A. Micchelli, M.~Pontil, and Y.~Ying.
\newblock Universal multi-task kernels.
\newblock \emph{The Journal of Machine Learning Research}, 9:\penalty0
  1615--1646, 2008.

\bibitem[Combettes and Pesquet(2011)]{Combettes2011}
P.~L. Combettes and J.-C. Pesquet.
\newblock Proximal splitting methods in signal processing.
\newblock In \emph{Fixed-point algorithms for inverse problems in science and
  engineering}, pages 185--212. Springer, 2011.

\bibitem[Cox(1961)]{Cox1961}
D.~R. Cox.
\newblock Tests of separate families of hypotheses.
\newblock \emph{Proc 4th Berkeley Symp Math Stat Probab}, 1:\penalty0 105--123,
  1961.

\bibitem[d'Alch{\'e} Buc et~al.(2012)d'Alch{\'e} Buc, Lim, Michailidis, and
  Senbabaoglu]{dalche2012}
Florence d'Alch{\'e} Buc, N{\'e}h{\'e}my Lim, George Michailidis, and Yasin
  Senbabaoglu.
\newblock {Estimation of nonparametric dynamical models within Reproducing
  Kernel Hilbert Spaces for network inference}.
\newblock In \emph{{Parameter Estimation for Dynamical Systems - PEDS II,
  Eindhoven, June 4-6, 2012}}, Eindhoven, Netherlands, June 2012. {Bart Bakker,
  Shota Gugushvili, Chris Klaassen, Aad van der Vaart}.
\newblock URL \url{https://hal.archives-ouvertes.fr/hal-01084145}.

\bibitem[Dinuzzo(2011)]{Dinuzzo2011}
Francesco Dinuzzo.
\newblock Learning functional dependencies with kernel methods.
\newblock \emph{Scientifica Acta}, 4\penalty0 (1):\penalty0 MS--16, 2011.

\bibitem[Dondelinger et~al.(2013)Dondelinger, Husmeier, Rogers, and
  Filippone]{Dondelinger2013}
F.~Dondelinger, D.~Husmeier, S.~Rogers, and M.~Filippone.
\newblock Ode parameter inference using adaptive gradient matching with
  gaussian processes.
\newblock In \emph{AISTATS}, volume~31 of \emph{JMLR Proceedings}, pages
  216--228. JMLR.org, 2013.

\bibitem[Ellner et~al.(2002)Ellner, Seifu, and Smith]{Ellner2002}
S.~P. Ellner, Y.~Seifu, and R.~H. Smith.
\newblock Fitting population dynamic models to time-series data by gradient
  matching.
\newblock \emph{Ecology}, 83\penalty0 (8):\penalty0 2256--2270, 2002.

\bibitem[Evgeniou and Pontil(2004)]{evgeniou2004regularized}
T.~Evgeniou and M.~Pontil.
\newblock Regularized multi--task learning.
\newblock In \emph{Proceedings of the tenth ACM SIGKDD international conference
  on Knowledge discovery and data mining}, pages 109--117. ACM, 2004.

\bibitem[Fitzhugh(1961)]{Fitzhugh1961}
R.~Fitzhugh.
\newblock Impulses and physiological states in theoretical models of nerve
  membrane.
\newblock \emph{Biophys J.}, 1\penalty0 (6):\penalty0 445--66, Jul; 1961.

\bibitem[Gugushvili and Klaassen(2011)]{Gugushvili2011}
S.~Gugushvili and C.A.J. Klaassen.
\newblock {Root-n-consistent parameter estimation for systems of ordinary
  differential equations: bypassing numerical integration via smoothing}.
\newblock \emph{Bernoulli}, to appear, 2011.

\bibitem[Ha~Quang and Bazzani(2013)]{Quang2013}
M.~Ha~Quang and V.~Bazzani, L.and~Murino.
\newblock A unifying framework for vector-valued manifold regularization and
  multi-view learning.
\newblock In \emph{Proc. of the 30th Int. Conf. on Machine Learning, {ICML}
  2013,}, pages 100--108, 2013.

\bibitem[Hirsch et~al.(2004)Hirsch, Smale, and Devaney]{Hirsch04}
M.~Hirsch, S.~Smale, and Devaney.
\newblock \emph{Differential Equations, Dynamical Systems, and an Introduction
  to Chaos (Edition: 2)}.
\newblock Elsevier Science \& Technology Books, 2004.

\bibitem[Kadri et~al.(2011)Kadri, Rabaoui, Preux, Duflos, and
  Rakotomamonjy]{Kadri2010}
H.~Kadri, A.~Rabaoui, P.~Preux, E.~Duflos, and A.~Rakotomamonjy.
\newblock Functional regularized least squares classiffication with
  operator-valued kernels.
\newblock In \emph{In Proc. International Conference on Machine Learning},
  2011.

\bibitem[Lim et~al.(2013)Lim, Senbabaoglu, Michailidis, and
  d'Alch{\'{e}}{-}Buc]{Lim2013a}
N.~Lim, Y.~Senbabaoglu, G.~Michailidis, and F.~d'Alch{\'{e}}{-}Buc.
\newblock Okvar-boost: a novel boosting algorithm to infer nonlinear dynamics
  and interactions in gene regulatory networks.
\newblock \emph{Bioinformatics}, 29\penalty0 (11):\penalty0 1416--1423, 2013.

\bibitem[Lim et~al.(2014)Lim, d'Alch{\'e} Buc, Auliac, and
  Michailidis]{Lim2013b}
N.~Lim, F.~d'Alch{\'e} Buc, C.~Auliac, and G.~Michailidis.
\newblock {Operator-valued Kernel-based Vector Autoregressive Models for
  Network Inference}.
\newblock hal-00872342, 2014.
\newblock URL \url{https://hal.archives-ouvertes.fr/hal-00872342}.

\bibitem[Micchelli and Pontil(2005)]{Micchelli2005}
C.~A. Micchelli and M.~Pontil.
\newblock On learning vector-valued functions.
\newblock \emph{Neural Computation}, 17\penalty0 (1):\penalty0 177--204, 2005.

\bibitem[Nagumo et~al.(1962)Nagumo, Arimoto, and Yoshizawa]{Nagumo1962}
J.~Nagumo, S~Arimoto, and S~Yoshizawa.
\newblock An active pulse transmission line simulating nerve axon.
\newblock \emph{Proceedings of the IRE}, 50\penalty0 (10):\penalty0 2061--2070,
  1962.

\bibitem[Pearce and Wand(2006)]{Pearce}
N.~D. Pearce and M.P. Wand.
\newblock Penalized splines and reproducing kernels.
\newblock \emph{The American Statistician}, 60\penalty0 (3), 2006.

\bibitem[Pedrick(1957)]{Pedrick57}
George~B Pedrick.
\newblock Theory of reproducing kernels for hilbert spaces of vector valued
  functions.
\newblock Technical report, DTIC Document, 1957.

\bibitem[Peifer and Timmer(2007)]{peifer07}
M~Peifer and J~Timmer.
\newblock Parameter estimation in ordinary differential equations for
  biochemical processes using the method of multiple shooting.
\newblock \emph{Systems Biology, IET}, 1\penalty0 (2):\penalty0 78--88, 2007.

\bibitem[Ramsay et~al.(2007)Ramsay, Hooker, Cao, and Campbell]{Ramsay2007}
J.O. Ramsay, G.~Hooker, J.~Cao, and D.~Campbell.
\newblock Parameter estimation for differential equations: A generalized
  smoothing approach.
\newblock \emph{Journal of the Royal Statistical Society (B)}, 69:\penalty0
  741--796, 2007.

\bibitem[Richard et~al.(2012)Richard, Savalle, and Vayatis]{Richard2012}
E.~Richard, P.-A. Savalle, and N.~Vayatis.
\newblock Estimation of simultaneously sparse and low rank matrices.
\newblock In John Langford and Joelle Pineau, editors, \emph{ICML-2012}, pages
  1351--1358, New York, NY, USA, July 2012. Omnipress.
\newblock ISBN 978-1-4503-1285-1.

\bibitem[Senkene and {Tempel'man}(1973)]{Senkene1973}
E.~Senkene and A.~{Tempel'man}.
\newblock Hilbert spaces of operator-valued functions.
\newblock \emph{Lithuanian Mathematical Journal}, 13\penalty0 (4):\penalty0
  665--670, 1973.

\bibitem[Varah(1982)]{Varah1982}
J.~M. Varah.
\newblock A spline least squares method for numerical parameter estimation in
  differential equations.
\newblock \emph{SIAM J.sci. Stat. Comput.}, 3\penalty0 (1):\penalty0 28--46,
  1982.

\bibitem[Vuong(1989)]{Vuong1989}
QH~Vuong.
\newblock Likelihood ratio tests for model selection and non-nested hypotheses.
\newblock \emph{Econometrica}, 57\penalty0 (2):\penalty0 307--333, 1989.
\newblock \doi{10.2307/1912557}.

\bibitem[Wahba(1990)]{Wahba90}
G.~Wahba.
\newblock \emph{Spline model for observational data}.
\newblock Philadelphia, Society for Industrial and Applied Mathematics, 1990.

\bibitem[Yuan and Lin(2006)]{yuan2006model}
M.~Yuan and Y.~Lin.
\newblock Model selection and estimation in regression with grouped variables.
\newblock \emph{J. of the Royal Statistical Society: Series B}, 68\penalty0
  (1):\penalty0 49--67, 2006.

\end{thebibliography}
\end{document}